\documentclass{article}

\PassOptionsToPackage{numbers}{natbib}
\usepackage[final]{neurips_2020}

\usepackage[utf8]{inputenc} % allow utf-8 input
\usepackage[T1]{fontenc}    % use 8-bit T1 fonts
\usepackage{hyperref}       % hyperlinks
\usepackage{url}            % simple URL typesetting
\usepackage{booktabs}       % professional-quality tables
\usepackage{amsfonts}       % blackboard math symbols

\usepackage{nicefrac}       % compact symbols for 1/2, etc.
\usepackage{microtype}      % microtypography

\usepackage{multirow}
\usepackage{graphicx}
\usepackage{subfigure}
\usepackage{xcolor}
\usepackage{tcolorbox}
\usepackage{enumitem}
\usepackage{array,makecell}
\usepackage{mwe} 
\usepackage{caption} 
\usepackage{makecell}
\usepackage{comment}

\definecolor{darkcyan}{rgb}{0.0, 0.55, 0.55}
\definecolor{darkgreen}{rgb}{0.0, 0.2, 0.13}

\newcommand*\rot{\rotatebox{90}}

\usepackage{amsmath}

\title{Understanding Anomaly Detection with Deep Invertible Networks through Hierarchies of Distributions and Features}

\author{%
  Robin Tibor Schirrmeister\!\thanks{This work was partially done during an internship at the Bosch Center for Artificial Intelligence.}\\
  University Medical Center Freiburg \\
  Bosch Center for Artificial Intelligence
\\
  \texttt{\small{robin.schirrmeister@uniklinik-freiburg.de}}\\
  \And
  Yuxuan Zhou\\
  Bosch Center for Artificial Intelligence\\
  \texttt{\small{Yuxuan.Zhou@bosch.com}}\\
  \And
  Tonio Ball\\
  University Medical Center Freiburg\\
   \texttt{\small{tonio.ball@uniklinik-freiburg.de}}
  \And
  Dan Zhang\\
  Bosch Center for Artificial Intelligence\\
  \texttt{\small{Dan.Zhang2@bosch.com}}
}

\begin{document}

\maketitle

%\vspace{-2em}
\begin{abstract}
Deep generative networks trained via maximum likelihood on a natural image dataset like CIFAR10 often assign high likelihoods to images from  datasets with different objects (e.g., SVHN). We refine previous investigations of this failure at anomaly detection for invertible generative networks and provide a clear explanation of it as a combination of model bias and domain prior: Convolutional networks learn similar low-level feature distributions when trained on any natural image dataset and these low-level features dominate the likelihood. Hence, when the discriminative features between inliers and outliers are on a high-level, e.g., object shapes, anomaly detection becomes particularly challenging. To remove the negative impact of model bias and domain prior on detecting high-level differences, we propose two methods, first, using the log likelihood ratios of two identical models, one trained on the in-distribution data (e.g., CIFAR10) and the other one on a more general distribution of images (e.g., 80 Million Tiny Images). We also derive a novel outlier loss for the in-distribution network on samples from the more general distribution to further improve the performance. Secondly, using a multi-scale model like Glow, we show that low-level features are mainly captured at early scales. Therefore, using only the likelihood contribution of the final scale performs remarkably well for detecting high-level feature differences of the out-of-distribution and the in-distribution. This method is especially useful if one does not have access to a suitable general distribution. Overall, our methods achieve strong anomaly detection performance in the unsupervised setting, and only slightly underperform state-of-the-art classifier-based methods in the supervised setting. Code can be found at \url{https://github.com/boschresearch/hierarchical_anomaly_detection}.

\end{abstract}
\section{Introduction}\label{sec:introduction}

One line of work for anomaly detection  - to detect if a given input is from the same distribution as the training data - uses the likelihoods provided by generative models. Through likelihood maximization, they are trained to yield high likelihoods on the in-distribution inputs (a.k.a. inliers).\footnote{
We note that likelihood is a function of the model given the data. As the model parameters are trained to model the data density, we may abuse likelihood and density in the paper for simplicity.} %Also note that, according to Shannon’s source coding theorem~\citep{Shannon1948}, likelihood maximization is equivalent to training the generative model to compress the inputs with a minimum expected encoding size in bits or bits per dimension (bps). Therefore, we will use likelihood and encoding size interchangeably throughout this paper. Shannon's source coding theorem applies for encoding discrete random variables and image data is naturally discrete. However, it is a common practice to instead work with continuous data and add dequantization noise to mirror the discrete case~\citep{Theis2015d}.
After training, one may  expect out-of-distribution inputs (a.k.a. outliers) to have lower likelihoods than the inliers. However, this is often not the case. %However, directly using the deep generative network's likelihoods often performs poorly at detecting images that contain the wrong  objects. 
For example,~\citet{nalisnick2018do} showed that generative models trained on CIFAR10 \citep{Cifar10_Krizhevsky09learningmultiple} assign higher likelihoods to SVHN~\citep{Netzer_SVHN} than to CIFAR10 images.

Several works have investigated a potential reason for this failure: The image likelihoods of deep generative networks can be well-predicted from simple factors. For example, the deep generative networks' image likelihoods highly correlate with: the image encoding sizes from a lossless compressor such as PNG \citep{Serra2020Input}; background statistics, e.g., the number of zeros in Fashion-MNIST/MNIST images \citep{ren_likelihood_ratio}; smoothness and size of the background \citep{krusinga2019understanding}. These factors do not directly correspond to the type of object, hence the type of object does not affect the likelihood much. %In short, the image likelihoods of deep generative networks can be well-predicted from simple factors not directly related to the object type. 

%\dan{also what is the relation to deep image prior paper}
In this work, we first synthesize these findings into the following hypothesis: A convolutional deep generative network trained on any image dataset learns low-level local feature relationships common to all images - such as smooth local patches - and these local features, forming the \emph{domain prior}, dominate the likelihood. One can therefore expect a smoother dataset like SVHN to have higher likelihoods than a less smooth one like CIFAR10, irrespective of the image dataset the network was trained on. Following prior works, we take Glow networks \citep{NIPS2018_8224} as the baseline model for our study.

%In this work, we first synthesize these findings into the following hypothesis: A convolutional deep generative network trained on \emph{any} image dataset learns low-level local feature relationships common to \emph{all} images (a domain prior) and these local features dominate the likelihood.  The prime example of these low-level feature relationships are smooth local patches. One can therefore expect a smoother dataset like SVHN to have higher likelihoods than a less smooth one like CIFAR10, irrespective of the image dataset the network was trained on. Following  prior works, we use invertible Glow networks \citep{NIPS2018_8224} as the baseline model for our study.

Next, we report several new findings to support the hypothesis: (1) Using a fully-connected instead of a convolutional Glow network, likelihood-based anomaly detection works much better for Fashion-MNIST vs. MNIST, indicating a convolutional model bias. (2)  Image likelihoods of Glow models trained on more general datasets, e.g., 80 Million Tiny Images (Tiny), have the highest average correlation with image likelihoods of models trained on other datasets, indicating a hierarchy of distributions from more general distributions (better for learning domain prior) to more specific distributions. (3) The likelihood contributions of the final scale of the Glow network correlate less between different Glow networks than the likelihood contributions of the earlier scales, while the overall likelihood is dominated by the earlier scales. This indicates a hierarchy of features inside the Glow network scales, from more generic low-level features that dominate the likelihood to more distribution-specific high-level features that are more informative about object categories.

Finally, leveraging the two novel views of a hierarchy of distributions and a hierarchy of features, we propose two likelihood-based anomaly detection methods. From the hierarchy-of-distributions view, we use likelihood ratios of two identical generative architectures (e.g., Glow), one trained on the in-distribution data (e.g., CIFAR10) and the other one on a more general distribution (e.g., 80 Million Tiny Images), refining previous likelihood-ratio-based methods. To further improve the performance, we additionally train our in-distribution model on samples from the general distribution using a novel outlier loss.
%, thus combining a likelihood-ratio-based approach with training on known outliers
From the hierarchy-of-features view, we show that using the likelihood contribution of the final scale of a multi-scale Glow model performs remarkably well for anomaly detection.

Our manuscript advances the understanding of the anomaly detection behavior of deep generative neural networks by synthesizing previous findings into two novel viewpoints accounting for the hierarchical nature of natural images. Based on this concept, we propose two new anomaly detection methods which reach strong performance, especially in the unsupervised setting. Our experiments are more extensive than previous likelihood-ratio-based methods on images, especially in the unsupervised setting, and therefore also fill an important empirical gap in the literature.

\section{Common Low-Level Features Dominate the Model Likelihood}\label{hypothesisr}
\begin{figure}
\begin{minipage}[b]{0.3\textwidth}
\centering
\includegraphics[width=\linewidth]{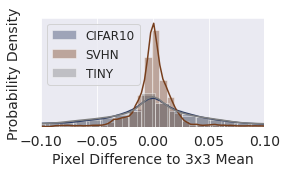}
\captionof*{figure}{\textbf{A}}
\end{minipage}
\hspace{0.02cm}
\begin{minipage}[b]{0.22\textwidth}
\centering
\includegraphics[width=\linewidth]{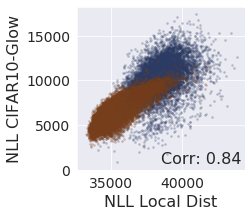}
  \captionof*{figure}{\textbf{B}}
\end{minipage}
\hspace{0.02cm}
\begin{minipage}[b]{0.4\textwidth}
\centering

\begin{tabular}{ccccc}
     \hline
     \textbf{\makecell{Conv Glow \\ trained on:}} & \rot{\textbf{Local Glow}} & \rot{\textbf{Dense Glow}} &
     \rot{\textbf{Conv Glow}} \\
    \hline
    \textbf{CIFAR10} &  1.0 & 0.86 & 1.0 \\
      \textbf{SVHN} & 0.96 &  0.90 & 0.97 \\
      \textbf{TINY} &  1.0 &  0.86 & 1.0 \\
     \hline
    \end{tabular}
  \captionof*{figure}{\textbf{C}}
\end{minipage}
\captionof{figure}{
Low-level features and model bias. \textbf{A}: Distributions over local pixel value differences have overlapping high-density regions for Tiny, SVHN, CIFAR10 . \textbf{B}: Likelihoods extracted from the local pixel-difference distributions correlate with CIFAR10-Glow likelihoods. \textbf{C}: Likelihood correlations for different types of Glow networks trained on CIFAR10 with regular convolutional Glow networks trained on CIFAR10, SVHN and Tiny. Correlations are almost the same for convolutional Glow networks and local Glow networks trained on $8\times 8$ patches. Correlations are smaller for fully-connected/dense Glow networks.
}
\label{fig:low-level-features}
\end{figure}

%on images of SVHN and CIFAR10. Local/Dense/Conv Glow all trained on CIFAR10.

The following hypotheses synthesized from the findings of prior work motivate our methods to detect if an image is from a different object recognition dataset:
\begin{enumerate}[topsep=0pt,parsep=1pt,leftmargin=0.6cm]
    \item Distributions of low-level features form a domain prior valid for all natural images (see Fig. \ref{fig:low-level-features}A).
    \item These low-level features contribute the most to the overall likelihood assigned by a deep generative network to any image (see Fig. \ref{fig:low-level-features}B and C).
    \item How strongly which type of features contributes to the likelihood is influenced by the model bias (e.g., for convolutional networks, local features dominate) (see Fig. \ref{fig:low-level-features}C).
\end{enumerate}

For the first hypothesis, we start from defining low-level features. They are features that can be extracted in few computational steps (these will be local features for convolutional models like Glow). As an example, we use the difference of a pixel value to the mean of its $3\times 3$ neighbouring pixels. %neighbours (smoother images will produce smaller differences) 
As natural images are smooth, the distributions over such per-pixel difference of SVHN, CIFAR10 and 80 Million Tiny Image (Tiny) depicted in Fig. \ref{fig:low-level-features}A are all zero centered. Smoother images will produce smaller differences among neighbouring pixels, therefore the density of SVHN has the highest peak around zero. A significant overlapping high-density regions for SVHN, CIFAR10 and Tiny show that these low-level features are common to natural images, and thus not useful for anomaly detection. %\dan{not sure we can generally say "low level features" are not useful, as this example only shows smoothness is not useful}
%The distribution over these per-pixel differences (see Fig. \ref{fig:low-level-features}A) have overlapping high-density regions for SVHN, CIFAR10 and 80 Million Tiny Images (Tiny), showing that these low-level features are not useful for anomaly detection. Smoother images will produce smaller differences, and thus SVHN 

Next, we examine the second hypothesis by showing that low-level per-image likelihoods highly correlate with Glow network likelihoods. %dan: here I also remove pseudo, as your 8x8 case is not pseudo.
%To obtain pseudo-likelihoods, 
On the pixel level, we compute the pixel difference and estimate its density using a histogram with $100$ equally distanced bins. Using the estimated density, we can get the conditional (on $3\times 3$ neighbours) per-pixel likelihoods of each image. Summing the per-pixel likelihoods over the entire image, we obtain pixel-level per-image pseudo-likelihood, which is not a correct likelihood of image, but a proxy measure of low-level feature contributions to the image-level likelihood. 
The low-level pseudo-likelihoods of SVHN and CIFAR10 images have Spearman correlations\footnote{All results in this section are qualitatively the same with Pearson correlations.} $> 0.83$ with likelihoods of Glow networks trained on CIFAR10 (see Fig \ref{fig:low-level-features}B), SVHN or Tiny. %To further check if low-level local features dominate the likelihood, %duplicated with "further suggesting ...."
We also trained small modified Glow networks on $8\times 8$ patches cropped from the original image. These local Glow networks' likelihoods correlate even more with the full Glow networks' likelihoods ($>0.95$), further suggesting low-level local features dominate the total likelihood (see Fig \ref{fig:low-level-features}C). To validate that the low-level features dominating the likelihoods are independent of the semantic content of the image, we mix two images in Fourier space by combining the amplitudes of one image's Fourier transform with the phases of the other image's Fourier transform, and then evaluate the likelihood of the resulting image using the same pretrained Glow model (see supp. material S3 for details). The mixed images are semantically much more coherent with the images that provide the phase information, yet their Glow likelihoods correlate more strongly with the Glow likelihoods of images sharing the same amplitudes ($> 0.8$ vs. $< 0.05$).

Lastly, we show that what features are extracted and how much each feature contributes to the likelihood depend on the type of model. When training a modified Glow network that uses fully connected instead of convolutional blocks (see supp. material S2.2) on Fashion-MNIST and MNIST, the image likelihoods among them do not correlate (Spearman correlation $-0.2$). The fully-connected Fashion-MNIST network achieves worse likelihoods ($4.8$ vs. $2.9$ bpd), but is much better at anomaly detection ($81\%$ AUROC for Fashion-MNIST vs. MNIST; $15\%$ AUROC for convolutional Glow).

Consistent with non-distribution-specific low-level features dominating the likelihoods, we find full Glow networks trained independently on CIFAR10, CIFAR100, SVHN or Tiny produce highly correlated image likelihoods (Spearman correlation $> 0.96$ for all pairs of models, see Fig. \ref{fig:low-level-features} \textbf{C}). The same is true to a lesser degree for Fashion-MNIST and MNIST (Spearman correlation $0.85$).

Taken together, the evidence suggests convolutional generative networks have model biases that guide them to learn low-level feature distributions of natural images (domain prior) well, at the expense of anomaly detection performance. Based on this understanding, next we propose two methods to remove this influence of model bias and domain prior on likelihood-based anomaly detection.

\section{Hierarchy of Distributions} \label{sec:hierarchy-distributions}

\begin{figure*}
    \centering
    \includegraphics[scale=0.45]{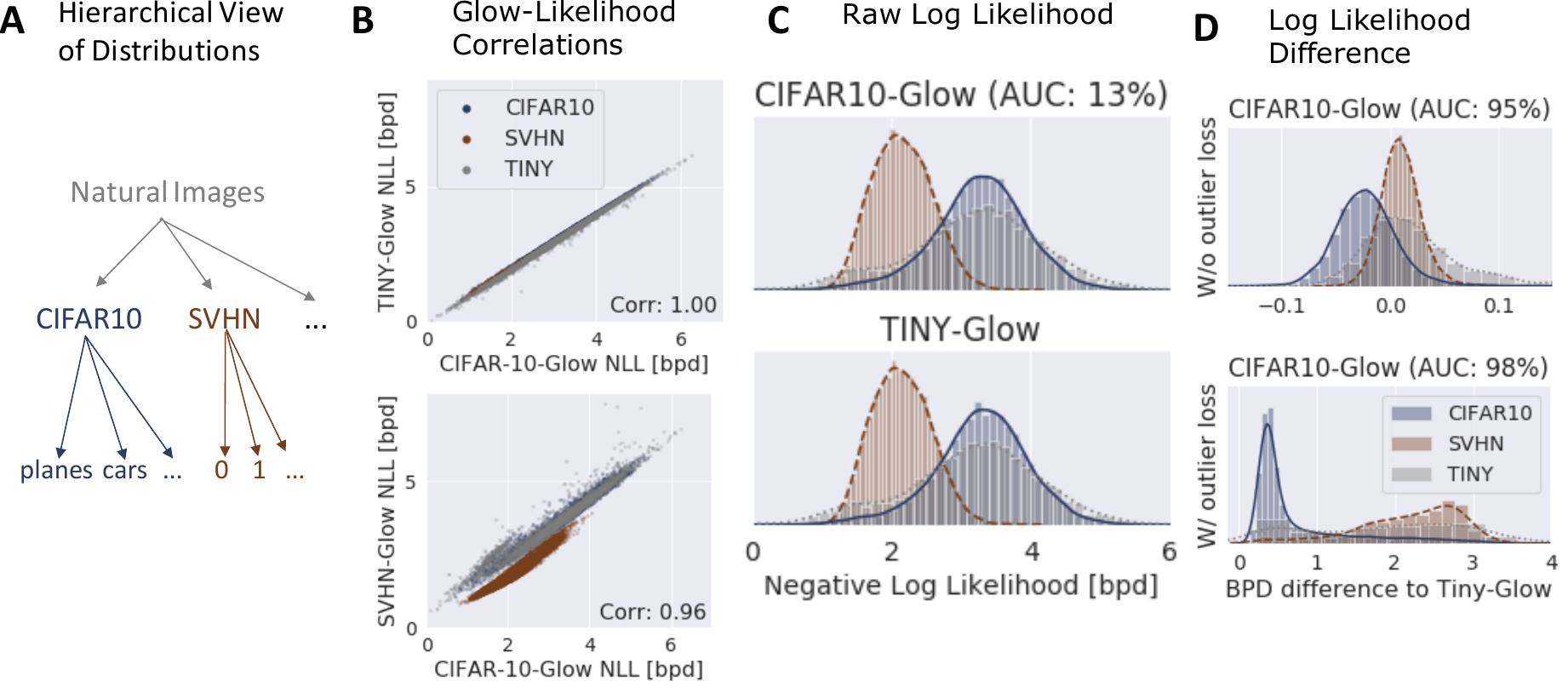}
    \caption{Overview of the hierarchy-of-distributions approach. \textbf{A}: Schematic hierarchical view of image distributions. To approximate the distribution of natural images, we use 80 Million Tiny images (Tiny). \textbf{B,C,D}: Results of Glow networks trained on CIFAR10, SVHN and Tiny. \textbf{B}: Likelihoods rank-correlate almost perfectly for the Glow networks trained on CIFAR10 and Tiny on all three datasets (top), while rank correlations remain very close to 1  for CIFAR10-Glow and SVHN-Glow (bottom), validating that the main likelihood contribution comes from the domain prior. \textbf{C}: Distribution plots show almost identical plots for CIFAR10 and Tiny-Glow and a low area under the receiver operating curve (AUC) for CIFAR10 vs. SVHN anomaly detection. \textbf{D}: In contrast, the log likelihood difference between CIFAR10-Glow and Tiny-Glow  reaches substantially higher AUCs (top), further increased by using our outlier loss (bottom) (see Section \ref{subsec:outlier-loss}).
    }
    \label{fig:anomaly-overview-distributions}
\end{figure*}

The models trained on Tiny have the highest average likelihood across all evaluated datasets. This inspired us to use a hierarchy of distributions: CIFAR10 and SVHN are subdistributions of natural images, CIFAR10-planes are a subdistribution of CIFAR10, etc. (see Fig. \ref{fig:anomaly-overview-distributions}).

We use this hierarchy of distributions to derive a log-likelihood-ratio-based anomaly detection method:

\begin{enumerate}[noitemsep,topsep=0pt,parsep=0pt,leftmargin=0.6cm]
    \item Train a generative network on a general image distribution like 80 Million Tiny Images
    \item Train another generative network on images drawn from the in-distribution, e.g., CIFAR10
    \item Use their likelihood ratio for anomaly detection
\end{enumerate}
%\dan{I think this way of describing the method costs too much space.}

Formally, given the general-distribution-network likelihood $p_g$ and the specific-in-distribution-network likelihood  $p_{in}$, our anomaly detection score (low scores indicate outliers) is:
\begin{equation}
\log\left(\frac{p_{\mathrm{in}}(x)}{p_{\mathrm{g}}(x)}\right)=\log\left(p_{\mathrm{in}}(x)) - \log(p_{\mathrm{g}}(x)\right).
\label{eq:likelihooddiff}
\end{equation}

%\dan{do we really need a subsection for this? I mean a paragraph would be enough right?}
\subsection{Outlier Loss}
\label{subsec:outlier-loss}
We also derive a novel outlier loss on samples $x_{\mathrm{g}}$ from the more general distribution based on the two networks likelihoods. Concretely, we use the log-likelihood ratio after temperature scaling by $T$ as the logit for binary classification:
\begin{equation}
L_{\mathrm{o}}=-\lambda\cdot\log\left(\sigma\left(\frac{\log(p_{\mathrm{g}}(x_{\mathrm{g}})) - \log(p_{\mathrm{in}}(x_{\mathrm{g}}))}{T}\right)\right)=-\lambda\cdot\log\left(\frac{\sqrt [T]{p_{\mathrm{g}}(x_{\mathrm{g}})}}{\sqrt[T]{p_{\mathrm{in}}(x_{\mathrm{g}})} + \sqrt[T]{p_{\mathrm{g}}(x_{\mathrm{g}})}}\right),
\label{eq:negative-loss}
\end{equation}
where $\sigma$ is the sigmoid function and $\lambda$ is a weighting factor.% See supp. material for pseudocode.

\subsection{Extension to the Supervised Setting}
In all previous parts, our method is presented in an unsupervised setting, where the labels of the inliers are unavailable. We extend our method  to the supervised setting with two main changes in our training. First, the Glow model $p_{\mathrm{in}}(x)$ uses a mixture of Gaussians for the latent $z$, i.e., each class corresponding to one mode. Second, the outlier loss is extended for each mode of $p_{\mathrm{in}}(x)$ by treating samples from the other classes as the negative samples, i.e., the same as $\{x_{\mathrm{g}}\}$ in Eq. \ref{eq:negative-loss}.

\section{Hierarchy of Features}\label{sec:hierarchy-features}
%\dan{shall we additionally pitch it as if general distribution is not there, this would be alternative. The existence of general distributions has been a common concern of the work in icml.}
\begin{figure*}
    \centering
    \includegraphics[width=\linewidth]{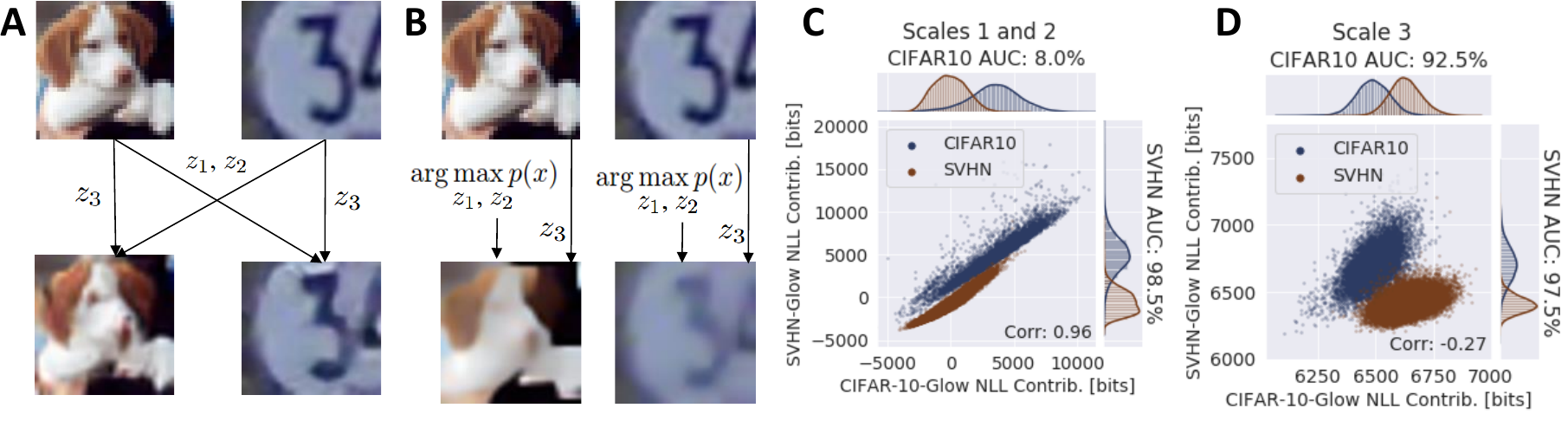}
    \caption{Overview of the hierarchy of features motivation. \textbf{A,B}: Showcasing features at different scales in the Glow-network.  Top images are examples from CIFAR10 and SVHN. \textbf{A}: Bottom two images are obtained by mixing features in the Glow-network trained on CIFAR10 as follows. We compute the three scale outputs of the Glow-network $z_1$, $z_2$ and $z_3$, mix them between both images and invert again. For the bottom-left image, we take the earlier-scale features $z_1$ and $z_2$ from the SVHN image and the final-scale features $z_3$  from the CIFAR10 image and vice versa for the bottom-right image. Note the image class is completely determined by  $z_3$. \textbf{B}: Images are optimized to maximize the CIFAR10-Glow-network likelihood $p(x)$ while keeping $z_3$ constant as follows. We keep $z_3$ from original Glow output fixed, and use the inverse pass of the network to optimize $z_1$ and $z_2$ via gradient ascent to maximize $\log(p(x))$. Only the global shape remains visible in the optimized images, while low-level structures have been blurred away. Such observation indicates that smoother local low-level features induce higher likelihood response of the model, once again confirming the strong influence of domain prior on the model likelhood. \textbf{C,D}: We use two Glow models trained on CIFAR10 and SVHN, respectively. The log likelihood they obtain on CIFAR10 and SVHN is split into log likelihood contributions of $z_1$, $z_2$ and $z_3$. The two plots show that (i) the summed contributions for $z_1$ and $z_2$ have very high rank correlation between both models (\textbf{C}); while the rank correlation drops for $z_3$ (\textbf{D}) and (ii) the range of the contributions is much larger for $z_1$ and $z_2$, showing that $z_1$ and $z_2$ dominate the total log likelihood.}
    \label{fig:anomaly-overview-features}
\end{figure*}

The image likelihood correlations between models trained on different datasets reduce substantially when evaluating the likelihood contributions of the final scale of the Glow network (see Fig. \ref{fig:anomaly-overview-features}). Here, the adopted Glow network has three scales. At the first two scales, i.e., $i=1$ and $2$, the layer output is split into two parts $h_i$ and $z_i$, where $h_i$ is passed onto the next scale and $z_i$ is output as one part of the latent code $z$.  The output at the last scale is $z_3$, which together with $z_1$ and $z_2$ makes up the complete latent code $z$. In terms of $y_1=(h_1,z_1)$, $y_2=(h_2,z_2)$, $y_3=z_3$ and $h_0=x$, the logarithm of the learned density $p(x)$  can be decomposed into per-scale likelihood contributions $c_i(x)$ as %\footnote{Note that, in contrast to other implementations, we do not condition $z_1$ on $z_2$ or $z_2$ on $z_3$. We discuss this choice further in the supplementary}
\begin{equation}
\log p(x) = \sum_i c_i(x) = \sum_i \log p_{\mathrm{z}}(z_i) + \log \left| \mathrm{det} \left(\frac{\partial y_i}{\partial h_{i-1}} \right) \right|\label{loglikelihoodscale}.
\end{equation}
The log-likelihood contributions $c_3(x)$ of the final third scale of Glow networks correlate substantially less than the full likelihoods for Glow networks trained on the different datasets ($0.26$ mean correlation vs. $0.99$ mean correlation). This is consistent with the observation that last-scale dimensions encode more global object-specific features (see Fig. \ref{fig:anomaly-overview-features} and \citep{dinh2016density}). Therefore, we use $c_3(x)$ as our anomaly detection score (low scores indicate outliers).

Note that, here we do not condition $z_1$ on $z_2$ or $z_2$ on $z_3$, whereas other implementations often make $z_1$ dependent of $z_2$ such as $z_1\sim N(f(z_2), g(z_2)^2)$ with $f,g$ being small neural networks. Such dependency is removable by transforming to $z_1'=\frac{(z_1 -f(z_2))}{g(z_2))}$, with $z_1'$ now independent of $z_2$ as $z_1'\sim N(0, 1)$, and this type of transformation can already be learned by an affine coupling layer applied to $z_1$ and $z_2$, hence the explicit conditioning of other implementations does not fundamentally change network expressiveness. We do not use it here and do not observe bits/dim differences between our implementation and those that use it (see supp. material S7.1 for details).
\section{Experiments}\label{sec:experiments}
%In this work, we use SVHN~\citep{Netzer_SVHN}, CIFAR10~\citep{Cifar10_Krizhevsky09learningmultiple}, CIFAR100~\citep{cifar100} as in-distribution and as OOD datasets for the anomaly detection evaluation.  Additionally, we add  LSUN~\citep{yu15lsun} as another OOD dataset. Results for further OOD datasets can be found in the supp. material. %
%For our general distribution dataset and our outlier loss dataset, we use 80 Million Tiny Images~\citep{80mtiny}. We take our hyperparameters from a public Glow repository\footnote{\url{https://github.com/y0ast/Glow-PyTorch/blob/master/train.py} (we do not use warmstart) and use our own Glow network implementation} such as Adamax as the optimizer (learning rate $5\cdot10^{-4}$, weight decay $5\cdot10^{-5}$) and 250 training epochs. For further details, we refer to the supp. material.

For the main experiments, we use SVHN~\citep{Netzer_SVHN}, CIFAR10~\citep{Cifar10_Krizhevsky09learningmultiple}, CIFAR100~\citep{cifar100} as inlier datasets  and use the same and LSUN~\citep{yu15lsun} as outlier datasets. Results for further outlier datasets can be found in the supp. material S7.4. 80 Million Tiny Images~\citep{80mtiny} serve as our general distribution dataset in the log likelihood-ratio based anomaly detection experiments and is also used in the outlier loss as given in Eq. \ref{eq:negative-loss} when training two generative models, i.e., Glow and PixelCNN++ (PCNN) (see supp. material S4 for their training details). In-distribution Glow and PixelCNN++ models are finetuned from the models pre-trained on Tiny for more rapid training, see supp. material S7.2 for details and ablation studies. Our reported results are averaged over 3 random seeds.%\dan{add our code link. if we add to abstract, will exceed the page limit.}

\subsection{Anomaly Detection based on Log-Likelihood Ratio}\label{subsec:logdiff-results}

\begin{table}%[htbp]
\setlength{\tabcolsep}{3pt}
\footnotesize
\caption{
Anomaly detection performance (AUCs in $\%$) of using the log-likelihood ratio of the in-distribution and general distribution model. %Top row lists two types of in-distribution model, i.e., Glow and PCNN.
For the general distribution models, we have three options, i.e., the general-purpose image compressor PNG~\citep{Serra2020Input} plus Tiny-Glow and Tiny-PCNN respectively trained on Tiny Images. Here, the Glow and PCNN trained on Tiny Images are also used as the starting point to train the in-distribution models. In the supp. material, we compare the results with training the two in-distribution models from scratch.}

\begin{center}
\begin{tabular}{cc|cccc|cccc}
 \hline
 &   \textbf{} & \multicolumn{4}{c|}{\textbf{Glow (in-dist.) diff to:}} &  
 \multicolumn{4}{c}{\textbf{PCNN (in-dist.) diff to:}}\\
 %\hline
 %\textbf{Gen-model}  & &  \textbf{None} & \textbf{PNG} & \textbf{Glow} & \textbf{PCNN} &  \textbf{None} & \textbf{PNG} & \textbf{Glow} & \textbf{PCNN} \\
\multirow{-2}{*}{\textbf{In-dist.}}&  \multirow{-2}{*}{\textbf{OOD}} & \textbf{None} & \textbf{PNG} & \textbf{Tiny-Glow} & \textbf{Tiny-PCNN} &  \textbf{None} & \textbf{PNG} & \textbf{Tiny-Glow} & \textbf{Tiny-PCNN} \\
\hline
\multirow{3}{*}{\textbf{SVHN}}
 & CIFAR10  & 98.3 & 74.4 & \textbf{100.0} & \textbf{100.0} & 97.9 & 76.8 & \textbf{100.0} & \textbf{100.0}\\
 & CIFAR100 & 97.9 & 79.5 & \textbf{100.0} & \textbf{100.0} & 97.4 & 81.3 & \textbf{100.0} & \textbf{100.0}\\
 & LSUN     & 99.6 & 96.8 & \textbf{100.0} & \textbf{100.0} & 99.4 & 98.1 & \textbf{100.0} & \textbf{100.0}\\
\hline
 %& Mean & 98.6 & 83.6 & \textbf{100.0} & \textbf{100.0} & 98.3 & 85.4 & \textbf{100.0} & \textbf{100.0}\\
%\hline
\multirow{3}{*}{\textbf{CIFAR10}}
 & SVHN     & 8.8 & 75.4 & \textbf{93.9} & 16.6 & 12.6 & 82.3 & \textbf{94.8} & 94.4\\
 & CIFAR100 & 51.7 & 57.3 & \textbf{66.8} & 53.4 & 51.7 & 57.1 & 57.5 & \textbf{63.5}\\
 & LSUN     & 69.3 & 83.6 & \textbf{89.2} & 16.8 & 74.8 & 87.6 & \textbf{93.6} & 92.9\\
\hline
 %& Mean & 43.3 & 72.1 & \textbf{83.3} & 29.0 & 46.4 & 75.7 & 82.0 & \textbf{83.6}\\
%\hline
\multirow{3}{*}{\textbf{CIFAR100}}
 & SVHN     & 10.3 & 68.4 & \textbf{87.4} & 18.3 & 13.7 & 76.4 & \textbf{91.3} & 90.0\\
 & CIFAR10  & 49.2 & 44.1 & 52.8 & \textbf{54.2} & 49.1 & 44.2 & 48.3 & \textbf{54.5}\\
 & LSUN     & 66.3 & 77.5 & \textbf{81.0} & 19.1 & 71.7 & 82.7 & \textbf{90.0} & 87.6\\
\hline
 %& Mean & 41.9 & 63.4 & \textbf{73.7} & 30.5 & 44.9 & 67.8 & 76.5 & \textbf{77.4}\\
%\hline
%\hline
 & Mean & 61.3 & 73.0 & \textbf{85.7} & 53.2 & 63.2 & 76.3 & 86.2 & \textbf{87.0} \\
\hline
\end{tabular}
\end{center}

\label{tab:base-specific-comparison-with-pixelcnn}\vspace{-0.4cm}
\end{table}
In Tab.~\ref{tab:base-specific-comparison-with-pixelcnn}, we compare the raw log-likelihood based anomaly detection (i.e., $\textbf{diff to: None}$) with the log-likelihood ratio based ones (i.e., $\textbf{diff to: PNG, Tiny-Glow, Tiny-PCNN}$). The raw log-likelihood based scheme underperforms the log-likelihood ratio based ones that use Tiny-Glow and Tiny-PCNN. However, when using PNG to remove the domain prior as proposed in~\cite{Serra2020Input}\footnote{The original results of \cite{Serra2020Input} are not comparable since they used the training set of the in-distribution for evaluation. We provide supplementary code that uses a publicly available CIFAR10 Glow-network with pretrained weights that roughly matches the results reported here.}, it sometimes performs worse than the raw log-likelihood based scheme, e.g., SVHN as the in-distribution vs. the other three OODs. This relates to the remaining model bias, as Glow trained on the in-distribution encodes the domain prior differently to PNG. Also note that using PCNN as the general-distribution model for Glow does not work well for CIFAR10/100. This is because Tiny-PCNN has very large bpd gains over Glow for the CIFAR-datasets and less large gains for SVHN. On average across datasets, it works best to use matching general and in-distribution models (Glow for Glow, PCNN for PCNN), validating our idea of a model bias. Also note that our SVHN vs. CIFAR10 results already outperform the likelihood-ratio-based results of \citet{ren_likelihood_ratio} slightly (93.9\%  vs. 93.1\% AUROC), and we observe further improvements with outlier loss in Section \ref{subsec:outlier-losses}. \citet{ren_likelihood_ratio} used a noised version of the in-distribution as the general distribution and only tested it on SVHN vs. CIFAR10. Comparing to Tiny, it is less representative as a domain prior, and thus its performance on more complicated datasets requires further assessment.

\paragraph{Medical Dataset} To further validate our log-likelihood ratio approach on a different domain, we setup an experiment on the medical BRATS Magnetic Resonance Imaging (MRI) dataset. We use one MRI modality as in-distribution and the other three as OOD. The raw likelihood, the log-likelihood ratio to Tiny-Glow and to BRATS-Glow (trained on all modalities) yield AUROC $53.3\%$, $68.3\%$ and $78.3\%$, respectively. So, Tiny also serves as a general distribution for the very different medical images, and a distribution from the more specific domain further improves the performance.

%\dan{Based on our rebuttal, I added some discussion on generalization of LLR and general distribution for other domain below.}

The log-likelihood ratio  approach can likely be applied to more than images. In the above, we have already shown the application  to typical image datasets and, without adaptation, to medical MRI images.  In the text/NLP domain, it may be used with Wikitext-2 as the general dataset, since Wikitext-2 already worked well as an outlier dataset in \citep{hendrycks2018deep}. In the audio domain, the domain prior may come from strong dependencies of the signal values on short timescales, similar to the smoothness of natural images. If a suitable general dataset needs to be created, it does not require labels and may even profit from noisy/unclean data. Therefore there is no principal obstacle preventing collection of such data, including concatenating existing datasets.

\subsection{Anomaly Detection based on Last-scale Log-likelihood Contribution}
As an alternative to remove the domain prior by using, e.g., Tiny-Glow, our hierarchy-of-features view suggests to use the log-likelihood contributed by the high-level features attained at the last scale of the Glow model. It is orthogonal to the log-likelihood ratio based scheme, and can be used when the general distribution is unavailable. As shown in~Tab. \ref{tab:lastscale}, using the raw log-likelihood on the last scale ($4\times4$), consistently outperforms the conventional log-likelihood comparison on the full scale, but performs slightly worse than using the log-likelihood ratio in the full scale. Note that we don't expect the log-likelihood ratio on the last scale to be the top performer, as the domain prior is mainly reflected by the earlier two scales, see Fig.~\ref{fig:anomaly-overview-features}.

\begin{table*}[t!]
    \begin{minipage}[t]{.49\linewidth}
     \caption{Raw log-likelihoods vs. log-likelihood ratios on different scales, where CIFAR10 is the in-distribution.}
      \centering
 \begin{tabular}{cccc}
             \hline
             \textbf{Out-dist}& \textbf{Scale} & \textbf{Raw} & \textbf{Diff}\\
            \hline
            \multirow{4}{*}{\textbf{SVHN}}
            
            & Full   & 8.8 & \textbf{93.9} \\
            & 16x16  & 7.0 & 84.6 \\
            & 8x8    & 13.5 & 48.9 \\
            & 4x4    & \textit{92.9} & 83.6 \\
             \hline
            \multirow{4}{*}{\textbf{CIFAR100}} 
             & Full   & 51.7 & \textbf{66.8} \\
             & 16x16  & 50.7 & 55.7 \\
             & 8x8    & 53.5 & 56.3 \\
             & 4x4    & \textit{60.0} & 66.1 \\
            \hline
            \multirow{4}{*}{\textbf{LSUN}}
             & Full   & 69.3 & \textbf{89.2} \\
             & 16x16  & 70.3 & 63.6 \\
             & 8x8    & 56.5 & 74.0 \\
             & 4x4    & \textit{82.8} & 75.1 \\
             \hline
            \end{tabular}\label{tab:lastscale}
    \end{minipage}%
    \hfill
    \begin{minipage}[t]{.49\linewidth}
      \centering
                \caption{Different outlier losses, where CIFAR10 is the in-distribution.}
        \begin{tabular}{ccccc}
             \hline
             \textbf{Out-dist}& \textbf{Loss} & \textbf{Raw}  & \textbf{4x4} &
             \textbf{Diff} \\
            \hline
            \multirow{3}{*}{\textbf{SVHN}} & None & 8.8 & 92.9 & 93.9 \\
             & Margin & 84.2 & 84.2 & 96.5  \\
             & Ours & 95.5 & 96.4 &  \textbf{98.6}   \\
             \hline
            \multirow{3}{*}{\textbf{CIFAR100}} & None & 51.7 & 60.0 & 66.8 \\
             & Margin & 72.3 & 71.7 & 71.1 \\
             & Ours & 84.9 & \textbf{85.4} & 84.5\\
             \hline
            \multirow{3}{*}{\textbf{LSUN}} & None & 69.3 & 82.8 & 89.2 \\
             & Margin &  82.0& 82.0 & 75.7 \\
             & Ours & 94.9  &\textbf{95.1} & 94.1   \\
             \hline
            \end{tabular}
            \label{tab:margin-ours}
    \end{minipage} 
\end{table*}

\subsection{Outlier Losses}
\label{subsec:outlier-losses}

% \begin{table}
% \setlength{\tabcolsep}{4pt}
% \caption{CIFAR10 with different outlier losses}
% \begin{center}
% \begin{tabular}{ccccc}
%  \hline
%  \textbf{Out-dist}& \textbf{Loss} & \textbf{Raw} \textbf{[Full]}  & \textbf{Raw} \textbf{[4x4]} &
%  \textbf{Diff} \\
% \hline
% \multirow{3}{*}{\textbf{SVHN}} & None & 12.6 & 91.1 & 95.1 \\
%  & Margin & 91.3 & 95.9 & 94.9  \\
%  & Ours & 94.8 & 95.4 &  \textbf{98.4}   \\
%  \hline
% \multirow{3}{*}{\textbf{CIFAR100}} & None & 51.2 & 59.7 & 66.1 \\
%  & Margin & 73.2 & 71.8 & 69.8 \\
%  & Ours & 83.7 & \textbf{84.4} & 83.5\\
%  \hline
% \multirow{3}{*}{\textbf{LSUN}} & None & 69.3 & 82.2 & 89.9 \\
%  & Margin &  84.6& 85.2 & 77.7 \\
%  & Ours & 93.8  &\textbf{94.1} & 92.8   \\
%  \hline
% \end{tabular}
% \label{tab:margin-ours}
% \end{center}\vspace{-0.5cm}
% \end{table}

When training the in-distribution model, we can use the images from Tiny as the outliers to improve the training. Tab.~\ref{tab:margin-ours} shows that our outlier loss as formulated in Eq. \ref{eq:negative-loss} consistently outperforms the margin loss~\citep{hendrycks2018deep} when combining with three different types of log-likelihood based schemes, i.e., raw log-likelihood, raw log-likelihood at the last scale $4\times 4$ and log-likelihood ratio. % Comparing with Tab.~\ref{tab:base-specific-comparison-with-pixelcnn} and Tab.~\ref{tab:lastscale}}, the proposed outlier loss further improves the raw log-likelihood at the last scale $4\times 4$ and log-likelihood ratio
%in combination with the log-likelihood ratio outperforms the margin loss~\citep{hendrycks2018deep} and never severely degrades when using the ratio instead of the raw log-likelihood, see Tab.~\ref{tab:margin-ours}. 
We note that as the margin loss leads to substantially less stable training than our loss, see the supp. material S7.5.

%\subsection{Tiny-Glow trained on in-dist as outlier}
We also experiment on adding the outlier loss to the training loss of Tiny-Glow, i.e., using the in-distribution samples as outliers. This further improves the anomaly detection performance, see \textbf{Diff}\dag~of Tab.~\ref{tab:overview-results}, while \textbf{Diff} only uses the outlier loss for training the in-distribution Glow-network. % 

\subsection{Unsupervised vs. Supervised Setting}
From the unsupervised to the supervised setting, Tab.~\ref{tab:overview-results} further reports the numbers achieved by using the class-conditional in-distribution Glow-network and treating inputs from other classes as outliers. We observe further improved anomaly detection performance. %This is despite the fact that due to resource constraints these models have only run half as many epochs as the other ones.

\label{ref:tiny-with-outlier-loss}
\begin{table*}%[htbp]
\caption{Anomaly detection performance summary (AUC in $\%$). The new term \textbf{Diff\dag} means to use in-distribution samples as the outliers to train Tiny-Glow, see Sec.~\ref{ref:tiny-with-outlier-loss}. \textbf{OE}, proposed by~\citet{hendrycks2018deep}, stands for margin-based outlier loss for PixelCNN++, \textbf{MSP-OE} from the same work stands for entropy of classifier predictions with entropy outlier loss.}

\begin{center}
\begin{tabular}{cc|cccc|cccccc}
 \hline
 \textbf{}& \textbf{Setting} & \multicolumn{4}{c}{\textbf{Unsupervised}}& \multicolumn{4}{c}{\textbf{Supervised}} \\
 \hline
 \textbf{In-dist} & \textbf{Out-dist} & \textbf{4x4} & \textbf{Diff} & \textbf{Diff}\dag & \textbf{OE} & \textbf{4x4} & \textbf{Diff} & \textbf{Diff}\dag & \textbf{MSP-OE}  \\
\hline
\multirow{3}{*}{\textbf{CIFAR10  }} & SVHN      & 96.4 & 98.6 & \textbf{99.0} & 75.8 & 96.1 & 98.6 & \textbf{99.1} & 98.4\\ 
                                    & CIFAR100  & 85.4 & 84.5 & \textbf{86.8} & 68.5 & 88.3 & 87.4 & 88.5 & \textbf{93.3}\\ 
                                    & LSUN      & 95.1 & 94.1 & \textbf{95.8} & 90.9 & 95.3 & 94.1 & 96.2 & \textbf{97.6}\\ 
\hline 
&Mean                                           & 92.3 & 92.4 & \textbf{93.8} & 78.4 & 93.3 & 93.4 & 94.6 & \textbf{96.4}\\ 
\hline 
\multirow{3}{*}{\textbf{CIFAR100 }} & SVHN      & 84.5 & 82.2 & \textbf{85.4} &  -   & \textbf{89.6} & 88.6 & 89.4 & 86.9\\ 
                                    & CIFAR10   & 61.9 & 59.8 & \textbf{62.5} &  -   & 67.0 & 64.9 & 65.3 & \textbf{75.7}\\ 
                                    & LSUN      & 84.6 & 82.4 & \textbf{85.4} &  -   & 85.7 & 84.3 & \textbf{86.3} & 83.4\\ 
\hline 
&Mean                                           & 77.0 & 74.8 & \textbf{77.8} &  -   & 80.8 & 79.3 & 80.3 & \textbf{82.0}\\ 
\hline 
\hline 
&Mean                                           & 84.7 & 83.6 & \textbf{85.8} &  -   & 87.0 & 86.3 & 87.5 & \textbf{89.2}\\ 
\hline 
\end{tabular}
\end{center}

\label{tab:overview-results}\vspace{-0.3cm}
\end{table*}

%\subsection{Comparison to related work}
Overall, our approach only slightly underperforms the approach MSP-OE~\citep{hendrycks2018deep} with inlier class labels (\textbf{Supervised}), while being substantially better without inlier class labels (\textbf{Unsupervised}), see Tab.~\ref{tab:overview-results}. In contrast to observations by \citet{hendrycks2018deep} for their unsupervised setup, we do not experience a severe degradation of the anomaly detection performance from the lack of class labels.

\begin{comment}
\subsection{Results for PixelCNN++}
While we focus on invertible Glow-networks in this study, we also confirmed that the hierarchy-of-distributions approach works for PixelCNN++. For in-distribution CIFAR10/CIFAR100 and SVHN, CIFAR10/100 and LSUN as OOD datasets, we retain very similar performance to Glow (80.4\% and 83.7\% mean AUC for the log-likelihood-difference, with and without outlier loss, respectively, whereas raw likelihoods only reach 45.7\%
\end{comment}

%\dan{I probably would put this analysis to supp. material and also medical dataset. }
\section{Related Work}
\label{sec:related-work}
We present an overview over anomaly detection approaches with a focus on recent work closely related to the ideas of a hierarchy of distributions and a hierarchy of features.

\textbf{Classifier-based Methods} Multi-class classifiers trained to discriminate in-distribution classes have been used for anomaly detection.~\citet{hendrycks17baseline} used maximum softmax response as the score of normality. %empirically showed multi-class classifiers tend to have smaller maximum softmax responses on outliers than inliers. %Therefore, one can exploit the statistics derived from softmax distributions for OOD detection. 
Different data augmentation schemes~\citep{liang2018enhancing, lee2018training,hendrycks2018deep,hendrycks2019selfsupervised} further enforced its performance. %Instead of using the softmax response, 
\citet{Leenips2018} alternatively modeled the class-conditional features attained by the hidden layers of the classifier as multivariate Gaussians, and then used the Mahalanobis distance of Gaussians for anomaly detection. Another recent work~\citep{grathwohl2020your} used the gradient norm of the log-sum-exp of the class logits over the input for anomaly detection. In the context of self-supervised learning, class becomes the type of transformations~\citep{bergman2020classification,golan2018deep}. Self-supervised contrastive training improved the anomaly detection performance of multi-class classifiers~\citep{winkens2020contrastive}.

%\textbf{Using a Classifier trained on the Inlier Data}  A variant to using the original class labels is to train a classifier on a self-supervised pretext task such as discriminating different geometric transformations applied to the inlier images \citep{bergman2020classification,golan2018deep}.%, namely, the former adding adversarial perturbations to the inliers and the latter resorting to a generative adversarial network for outlier synthesis. %, namely, adding small adversarial perturbations to the classifier inputs during training, and also by temperature scaling the softmax function. 
%In~\citep{lee2018training}, a GAN was concurrently trained with the classifier to synthesize OOD samples that are close to in-distribution ones. The classifier is then trained to output uniform predictive distribution on such synthesized OOD samples. 
%\citeauthor{hendrycks2018deep} proposed to use dataset that is free to have as the OOD samples to fine-tune the pre-trained classifier, i.e, exposing the classifier to outliers. 
%The recent work by  showed that anomaly detection can benefit from self-supervised training techniques, such as rotating the input image in one of four different degrees and training the network to predict the corresponding rotations.

Instead of exploiting multi-class classifiers, %
%\textbf{Using an Inlier-vs-Outlier Classifier}
a different approach is to train a one-class classifier to directly discriminate inliers and outliers. % %function to yield positive values on inliers and negative values on outliers. 
One-class support vector machines are trained to return positive values only in a small region containing the inliers and negative values elsewhere~\citep{scholkopf2000support}. This approach has also been used in forming the latent space of deep autoencoders~\citep{ruff2018deep,ruff2019deep, ruff2020rethinking}. \citet{ruff2020rethinking} also used samples from a general distribution as outliers. \citet{steinwart2005classification} drawn outliers from uniform distribution. In the supp. material S9, we also report results using a in-distribution-vs-general-distribution classifier.  %Another variant are methods that explicitly contrast inliers with an artificial distribution like the uniform distribution~. 

\textbf{Reconstruction-based Methods}
%\textbf{Using Reconstruction Errors} 
Another line of work is to learn the features and generation of inliers by reconstructing the training samples either in their input space or latent space, e.g.,~\citep{Pidhorskyi2018,Perera2019OCGANON,abati2019latent,kim2019rapp}. At test time, an outlier is then detected if reconstruction is poor. However, owing to large capacity of deep neural networks, reconstruction loss alone may not be a reliable metric for anomaly detection. \citet{huang2019out} proposed to additionally use the joint likelihood of latent variables, which is obtained by using a neural rendering model to invert multi-class CNN-based classifier.%\dan{if they use all latent variables, why this relates to last scale feature?} \robin{their result is in contradiction with us in my view. they even claim: "We contrast the distribution [all layer latents] to the distribution of p(z)observed in Glow by Nalisnick et al. (2018) as shown in Figure 1.  The z in the p(z)of Glow corresponds to the last layer of latent variables.  In Glow’s p(z)distribution, we observe some separation between the peaks of in-distribution images (CIFAR-10) and OoD images (SVHN) but also observe similar amount of separation between the peaks of train and test set of the in-distribution images.  This separation suggests that the final layer of latent variables captures details which are specific to the training set instead of general features of the whole data distribution. In contrast, the joint distribution [all layer latents] does not suffer from this problem because it uses information from the latent variables across all layers, causing it to not be overly sensitive to features only in the training set.Thus, [all layer latents] is a better distinguisher between in-distribution and OoD samples."} 
%  
%Another line of work uses autoencoder-based representation learning and generative modelling, e.g.,. Briefly, an autoencoder consists of a encoder-decoder pair. The encoder maps the input sample into a latent code that resides in a constrained latent space. As a compact representation of the input, the latent code is fed into a decoder to reconstruct the input. For anomaly detection, one expects the inliers to have much lower reconstruction errors than the outliers. Autoencoder-based anomaly detection can greatly benefit from techniques that improve representation learning of the encoder and generative modelling of the decoder.

%\cite{kim2019rapp}
%Among the generative modelling methods, variational autoencoders and invertible networks explicitly model the data distribution underlying the in distribution training samples. At the test time, whenever the inputs have their densities below a threshold, they are detected as out-of-distribution samples.

%\citep{Choi2020Novelty} augement the training of VQ-VAE with image blurring.

%autoencoder based approaches: \citep{Pidhorskyi2018,Perera2019OCGANON,abati2019latent}

%\dan{I wonder if the three papers are better grouped under auto-encoder based representation learning, as they worked due to some smart manipulation on the latent space, \ref{abati2019latent} also latent prior in addition to reconstruction error.}

\textbf{Input Likelihood-based Methods}
%Multi-class classifier attempts to learn the a-posteriori distribution $p(y\vert x)$ of the label $y$ given the input $x$. One important premise for $p(y\vert x)$ being defined is that $x$ must have a positive density, i.e., $p(x)>0$. Therefore, at the test time, it is important to check before classification if every received $x$ has a large enough density. 
Generative modeling through maximum likelihood estimation tries to enforce high likelihoods on inliers. Under the normalization constraint, the likelihoods of outliers are expected to be low (ideally zero). However, their anomaly detection performance is often unsatisfactory~\citep{Shafaei2018,nalisnick2018do,hendrycks2018deep}. Outliers may attain even higher likelihoods than inliers. %
Recent work~\citep{Nalisnick2019DetectingOI} related the poor performance to sampling in a high-dimensional space, namely, inliers being mapped to the typical set of the latent code rather than the high likelihood area. They proposed to address this issue by batch-wise anomaly detection, whose application is more limited than instance-wise anomaly detection. %Empirically, energy-based models have achieved better anomaly detection performance \citep{zhai2016deep}. 
A different approach \citep{che2019deepverifier} combined input likelihoods with inlier classifiers. \citet{che2019deepverifier} trained a class-conditional generative model with an auxiliary adversarial loss to disentangle the class information from the rest latent representation. The achieved performance is better than ours in the supervised setting, while our methods mainly target and work in the unsupervised setting. It can be interesting to exploit their way of incorporating the label information into our model training. %improve our methods in the supervised setting by exploiting their way of incorporating the label information. %\dan{but they did not suggested look at the last scale. to me, it is related work in the sense of building up the view.} \robin{Then move this sentence to its own paragraph? or where to move it? want to keep it.}\dan{maybe after [23] as, both [23] and it are to understand why ood failed}

Our work is also input likelihood-based. Our analysis in Section~\ref{hypothesisr} showed that convolutional networks trained on one natural image dataset will learn low-level feature distribution that is common to the whole domain, and such domain prior dominates the likelihood. The concurrent work~\citep{kirichenko2020why} found results consistent with ours. We further exploited hierarchies of distributions or hierarchies of features as explained in Section \ref{sec:hierarchy-distributions} and \ref{sec:hierarchy-features} to improve the anomaly detection performance.

\emph{Hierarchy of Distributions}: Our hierarchy-of-distributions likelihood-ratio method relates to prior \citep{ren_likelihood_ratio} and concurrent \citep{Serra2020Input} methods as follows. \citet{ren_likelihood_ratio} used a noised version of the in-distribution as the general distribution. Their method always requires training of two models for each in-distribution, our method has the option of only require one training of a general distribution model which we can reuse for any new in-distribution as long as it is as a subdistribution of the general distribution. In our method, the challenge is to find a suitable general distribution, while in their method the challenge is to find a suitable noise model. In the only rgb-image-setting they evaluated in \citep{ren_likelihood_ratio}, we show improved results over theirs, see Section \ref{subsec:logdiff-results}. \citet{Serra2020Input} used a generic lossless compressor such as PNG as their general distribution model and unfortunately only reported results on the training set of the in-distribution, making their results incomparable with any other works. We show improved performance over a reimplementation (see code repository) in Section \ref{subsec:logdiff-results}. Note that both \citep{ren_likelihood_ratio} and \citep{Serra2020Input} did not evaluate the cases where raw likelihoods work well, even though using the likelihood ratio may decrease performance in that case as seen in Section \ref{subsec:logdiff-results}.

\emph{Outlier Loss}: \citet{hendrycks2018deep} used a margin loss on images from a known outlier distribution (Tiny without CIFAR-images). We develop an outlier loss to improve the performance of likelihood models. It can be viewed as a combination of their idea of an outlier loss with our view of a hierarchy of distributions, and achieved an improved performance in the unsupervised setting in Section \ref{subsec:outlier-losses}.

\emph{Hierarchy of Features}: Regarding hierarchy of features, the closest related work investigated deep variational autoencoders that use a hierarchy of stochastic variables and found that later stochastic variables perform better at anomaly detection \citep{maaloe2019biva}. Furthermore, \citet{krusinga2019understanding} developed a method to  approximate probability densities from generative adversarial networks. The log densities are the sum of a change-of-volume determinant and a latent prior probability density. The latent log densities better reflected semantic similarity to the in-distribution than the full log density, however, no comparable anomaly detection results were reported. \citet{nalisnick2018do} also found the same for latent vs. full log densities of Glow networks, but also did not report anomaly detection results and did not look at individual scales of Glow networks.

%One recent state-of-the art purely supervised approach that uses a classifier, adversarial training and generative models is Deep Verifier Networks from~\citet{che2019deepverifier}. This approach outperforms the results reported here, showing that an approach using the class labels without an auxiliary outlier dataset can outperform using 80 Million Tiny Images as outliers. There are two potential reasons for the improved performance. First, they use an adversarial approach, that forces the class-conditional generative model to make its encodings independent of the class label, ensuring the high-level class-discriminative features. They report the importance of this part, consistent with our hypotheses. Second, there are still too many images in Tiny that would be valid CIFAR-images, especially valid images for the people classes in CIFAR100. Therefore, finding a more suitable outlier dataset as well as further improved objective functions are a promising future direction, for both the supervised and unsupervised settings.

\section{Discussion and Conclusion}
%Overall, our proposed approaches outperform other proposed approaches in the unsupervised setting and are competitive in the supervised setting.

In this work, we proposed two log-likelihood based metrics for anomaly detection, outperforming state of the art methods in the unsupervised setting and only slightly underperforming classifier-based methods in the supervised setting. For good anomaly detection performance from raw likelihoods, an additional loss (such as our outlier loss) that forces the model to assign low likelihoods for images with OOD-high-level features, e.g. wrong objects, is particularly beneficial according to empirical results. Our analysis points to a potential reason, namely that without an outlier loss, the likelihoods are almost fully determined by low-level features such as smoothness or the dominant color in an image. As such low level features are common to natural image datasets, they form a strong domain prior, presenting a difficult task to detect high-level differences between inliers and outliers, e.g., object classes.

%While the actual CIFAR-images cover less than 0.2\% of the 80 Million Tiny Images, many more images that would be valid CIFAR-images\citep{Cifar10_Krizhevsky09learningmultiple}.

An interesting future direction is how to best combine the hierarchy-of-distributions and hierarchy-of-features views into a single approach. Preliminary experiments freezing the first two scales of the Tiny-Glow model and only finetuning the last scale on the in-distribution have shown promise, awaiting further evaluation.

In summary, our approach shows strong anomaly detection performance particularly in the more challenging unsupervised setting and also allows a better understanding of generative-model-based anomaly detection by leveraging hierarchical views of distributions and features.

\section*{Broader Impact}
A better understanding of deep generative networks with regards to anomaly detection can help the machine learning research community in multiple ways. It allows to estimate which tasks deep generative networks may be suitable or unsuitable for when trained via maximum likelihood. With regards to that, our work helps more precisely understand the outcomes of maximum likelihood training. This more precise understanding can also help guide the design of training regimes that combine maximum likelihood training with other objectives depending on the task, if the task is unlikely to be solved by maximum likelihood training alone.

Anomaly detection in general itself has positive uses. For example, detecting anomalies in medical data can detect existing and developing medical problems earlier.  Safety of machine learning systems in healthcare, autonomous driving, etc., can be improved by detecting if they are processing data that is unlike their training distribution. 

Negative uses and consequences of anomaly detection can be that it allows tighter control of people by those with access to large computer and data, as they can more easily find unusual patterns deviating from the norm.  For example, it may also allow health insurance companies to detect unusual behavioral patterns and associate them with higher insurance costs. Similarly  repressive governments may detect unusual behavioral patterns to target tighter surveillance.

These developments may be steered in a better direction by a better public understanding and regulation for what purpose anomaly-detection machine-learning systems are developed and used.

\begin{ack}
This work was partially done during an internship of Robin Tibor Schirrmeister at the Bosch Center for Artificial Intelligence. A part of this work was supported by the german Federal Ministry of Education and Research (BMBF, grant RenormalizedFlows 01IS19077C). 

Yuxuan Zhou wants to thank Zhongyu Lou and Duc Tam Nguyen for the helpful discussions about the preliminary experiment results.

Robin Tibor Schirrmeister wants to thank Jan Hendrik Metzen, Polina Kirichenko, Pavel Izmailov, Manuel Watter and Dengfeng Huang for discussions and support.

\end{ack}
\newcommand{\beginsupplement}{%
	\setcounter{table}{0}
	\renewcommand{\thetable}{S\arabic{table}}%
	\setcounter{figure}{0}
	\renewcommand{\thefigure}{S\arabic{figure}}%
	\setcounter{section}{0}
	\renewcommand{\thepage}{S\arabic{page}} 
	\renewcommand{\thesection}{S\arabic{section}}  
	\setcounter{equation}{0}
	\renewcommand{\theequation}{S\arabic{equation}}
}
\bibliographystyle{plainnat}
\bibliography{references}
\clearpage
\section*{Supplementary outline}
\appendix
\beginsupplement
This document completes the presentation of the main paper with the following:

\begin{enumerate}[label=\textsc{S\arabic*:},leftmargin=1cm]
    \item Details about Glow and PixelCNN++ architectures, and the likelihood decomposition equation Eq. (3) of Sec.~4 in the main paper;% in Section \ref{sec:architecture}
    \item Details about the modified (local/fully connected) Glow architectures for the analysis in Sec.~2 of the main paper; %manuscript in Section \ref{sec:hypothesis-architectures} here
    \item Fourier-based analysis of influence of amplitude and phase  on likelihoods; %\ref{sec:fft-amplitude-phase}
    \item Details about training and evaluation of Glow and PixelCNN++, including hyperparameter choices and computing infrastructure; % in Section  \ref{sec:training-evaluation}
    \item Details on the used datasets and dataset splits; % in Section \ref{sec:datasets}
    \item Reasons why the results of \citet{Serra2020Input} are not comparable as is, and details of our reimplementation; % in Section \ref{sec:replicate-serra} 
    \item Further quantitative results, % in Section \ref{sec:quantitative-results},
    including maximum-likelihood performance (\ref{sec:bpds}), finetuning vs. from-scratch training (\ref{subsec:finetuning}), variance over seeds (\ref{subsec:seed-variance}), further outlier datasets (\ref{subsec:more-ood}) and different outlier losses (\ref{subsec:loss-types});
    \item Qualitative analysis of the different anomaly detection metrics; % in  quantitative results in Section  \ref{subsec:loss-types}
    \item Generative vs. discriminative approach for anomaly detection. %Qualitative results showing the effect of different outlier losses in Section \ref{sec:qualitative-results}
\end{enumerate}

Please also note the attached supplementary codes.

\section{Glow and PixelCNN++ architectures}\label{sec:architecture}

\subsection{Glow Network architecture}\label{subsec:full-glow-architecture}
Our implementation of the Glow network \citep{NIPS2018_8224} is based on a publicly available Glow implementation \footnote{\url{https://github.com/y0ast/Glow-PyTorch/}} with one modification explained in \ref{subsec:independent-z}. The multi-scale Glow network consists of three sequential scales processing representations of size $12\times 16\times 16$, $24\times 8\times 8$ and $48\times 4\times 4$ (channel $\times $ width $\times$ height). Each scale consists of a repeating sequence of activation normalization, invertible $1\times 1$ convolution and affine coupling blocks, see the original paper \citep{NIPS2018_8224} for details. Our Glow network, consistent with aforementioned public implementation, uses $32$ actnorm-$1\time 1$ conv-affine sequences per scale.

\subsection{PixelCNN++ architecture}
We use a publicly available PixelCNN++ implementation \footnote{\label{foot:pixelcnn++}\url{https://github.com/pclucas14/pixel-cnn-pp/tree/16c8b2fb8f53e838d705105751e3c56536f3968a}} with only a single change. We reduce the number of filters used across the model from $160$ to $120$ for fast single-GPU training.

\subsection{Independent $z_1$, $z_2$, $z_3$}\label{subsec:independent-z}
%\dan{maybe add some context, for instance, this is about which section in the main paper.}
For a multi-scale model like Glow, the overall likelihood consists of the contributions from different scales, see Eq. (3) in the main paper. In contrast to other implementations, we do not condition $z_1$ on $z_2$ or $z_2$ on $z_3$ in our Glow model as described in Sec.~\ref{subsec:full-glow-architecture}. Recall that Glow splits the complete latent code into per-scales latent codes $z_1$, $z_2$, $z_3$. Here, $z_1$ is the half of the output of the first scale that is not processed further. Many implementations make $z_1$ dependent of $z_2$ (and same for $z_2$ and $z_3$) as $z_1\sim N(f(z_2), g(z_2)^2)$ with $f,g$ being small neural networks. For ease of implementation, we do not do that, instead we directly evaluate $z_1$ under a standard-normal gaussian, so  $z_1\sim N(0, 1)$.

Note that this does not fundamentally alter network expressiveness. An affine coupling layer can already implement the same computation achieved by $z_1\sim N(f(z_2), g(z_2)^2)$. Imagine $z$ is split for the affine coupling layer into $z_1$ and $z_2$, with a coefficient network on $z_2$ used to compute the affine scale and translation coefficients $s, t$ to transform $z_1'=z_1\odot s(z_2) + t(z_2)$. Then if $s(z_2)= -f(z_2)$ and $t(z_2)= \frac{1}{g(z_2)}$ and $z_1\sim N(f(z_2), g(z_2)^2)$, it follows that $z_1'\sim N(0, 1)$. In other words, computing the mean and standard deviation for $z_1$ from $z_2$ is the same as normalizing $z_1$ by subtracting the mean and dividing by the standard deviation computed from $z_2$, which a regular affine coupling block can already learn.

In practice, there could still be differences due to the additional parameters, different kind of blocks used to implement $f$ and $g$, and the difference of computing the (log)$std$ or its inverse. However, we observe no appreciable bits/dim differences between our implementation and those using the explicit conditioning step, see Section \ref{sec:bpds}. 
%\dan{sounds a bit uncertain. You first say "Practically" and then say "In practice". Can we just keep the last sentence? Maybe add numbers here in addition to re-directing to Sec. 6}

%\dan{for the two subsections. they are both for section 2 in the main paper, maybe coud have a subsection: Experiment details for Sec. 2 and then they are explained that as two paragraphs. So far, I feel, it is a bit difficult for people to navigate the supp. material and see the link with the main paper.}

\section{Local and Fully Connected Architectures}\label{sec:hypothesis-architectures}

\subsection{Local Patches}\label{subsec:local-patches}
We designed our local patches experiment to train compact Glow-like models that can only process information from $8\times 8$ patches in the original datasets. Full-sized Glow networks process the full image using three scales as written in Section \ref{subsec:full-glow-architecture}. Our local Glow network instead processes local $8\times 8$ patches using a single scale. The $32\times 32$ input image is first cropped into $16$ non-overlapping $8\times 8$ patches. These $8\times 8$ patches are then processed independently by a local Glow network corresponding to a single scale of the full Glow network. In other words, we treat the image as if it consists of independent $8 \times 8$ patches. Evaluating the likelihoods of these patches, their sum is the likelihood of the image assigned by the local Glow model. Note that we aimed to create a network restricted to learn a general local domain prior and not one with the best maximum-likelihood performance.

\subsection{Fully Connected}\label{subsec:fully-connected}
We designed our fully-connected experiment to train fully-connected Glow networks that have a different model bias to regular convolutional Glow networks. We kept the three-scale architecture of Glow including the invertible subsampling steps at the beginning of each scale. Within each scale, the fully-connected Glow first flattens the representation, e.g. from a $12\times 16\times 16$ tensor per rgb-image to a $3072$-sized vector in the scale $1$. This vector is then processed by the usual sequence of actnorm-$1\times 1$-affine. We next detail the processing at the scale $1$, whereas the other scales follow the same design pattern. Activation normalization now processes $3072$ dimensions, so has substantially more parameters. The $1\times 1$ is now an invertible linear projection keeping the dimensionality, so a projection from $3072$ dimensions to $3072$ dimensions. To ensure training stability, we did not train the $1\times 1$-projections, but kept their parameters in the randomly initialized starting state. The fully-connected affine coupling block uses a sequence of linear layer ($1536\times512$) - ReLU - linear layer ($512\times512$) - ReLU - linear layer ($512\times 3072$) modules to compute the $1536$ translation and $1536$ scale coefficients. To allow fast single-GPU training, we reduced the number of actnorm-$1\times 1$-affine sequences from $32$ to $8$ per scale. Similar to Section \ref{subsec:local-patches}, the fully-connected network was designed to highlight the influence different model biases and not to reach the best maximum-likelihood performance.

\section{Fourier-based Amplitude/Phase Analysis}\label{sec:fft-amplitude-phase}

\begin{figure*}
    \centering
    \includegraphics[width=\linewidth]{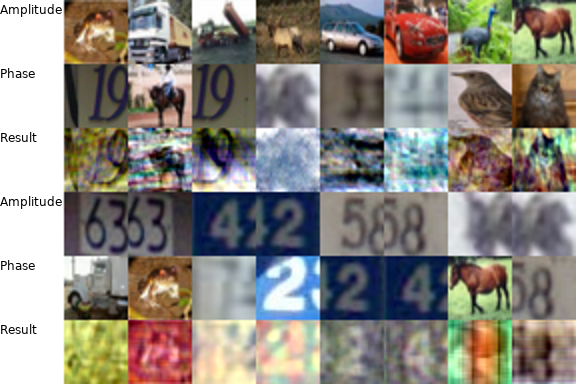}
    \caption{Image Mixup in frequency domain via Fourier transform. Fourier amplitudes taken from one image, phases from another and then inverted back to input space (Result). Note semantic content is more similar to phase image than to amplitude image.
    }
    \label{fig:fft-amplitude-phase}
\end{figure*}

To validate that low-level features dominate the likelihoods independent of the semantic content of the image, we create mixed images in Fourier space. Concretely, we:

\begin{enumerate}
    \item Compute the amplitudes and phases of a batch Fourier transformed images;
    \item Mix up images in their frequency domain by using one image' amplitudes and phases of the other image
    \item Apply the inverse Fourier transformation to invert these mixed images to the input domain 
\end{enumerate}

We show examples in Figure \ref{fig:fft-amplitude-phase}. Note that the mixed images are semantically much more similar to the image the phases were extracted from. We then compare the CIFAR10-Glow likelihoods on the original images and the mixed images for SVHN and CIFAR10 images. The likelihoods of the mixed images correlate much more with the amplitude-image likelihoods (Spearman correlation $>0.8$) than with the phase-image likelihoods (Spearman correlation $<0.05$).
\section{Training and Evaluation}\label{sec:training-evaluation}
\subsection{Glow Training}
We stayed close to the training setting of a publicly available Glow repository\footnote{\url{https://github.com/y0ast/Glow-PyTorch/blob/master/train.py} (we do not use warmstart)}. Namely, we use Adamax as the optimizer (learning rate $5\cdot10^{-4}$, weight decay $5\cdot10^{-5}$) and $250$ training epochs. The training setting also includes data augmentation (translations and horizontal flipping for CIFAR10/100, only translations for SVHN, Fashion-MNIST and MNIST). These settings are the same for all experiments (from-scratch training, finetuning, with and without outlier loss, unsupervised/supervised). On 80 Million Tiny Images, we use substantially less training epochs, so that the number of batch updates is identical between Tiny and the other experiments. All datasets were preprocessed to be in the range  $[-0.5,0.5 - \frac{1}{256}]$ as is standard practice for Glow-network training, see also supplementary code.

\subsection{PixelCNN++ Training}
We stayed close to the training setting of the public PixelCNN++ repository\footnote{\url{https://github.com/pclucas14/pixel-cnn-pp/tree/16c8b2fb8f53e838d705105751e3c56536f3968a}}. Namely, we use Adam as the optimizer (learning rate $2\cdot10^{-4}$, no weight decay, negligible learning rate decay $5\cdot10^{-6}$ every epoch). We substantially reduced the number of epochs from $5000$ to $120$ to save GPU resources, and since our aim here was to show general applicability of our methods to another type of model and not to reach maximum possible performance. Consistent with the public implementation, no data augmentation is performed.

\subsection{Numerical stabilization of the training} \label{subsec:numerical}
When training networks with outlier loss, negative infinities can appear for numerical reasons. This is actually expected as true outliers should ideally have likelihood zero and therefore log likelihood equal to negative infinity. These do not contribute to the loss in theory, but cause numerical issues. To ensure numerical stability during training, we remove examples that get assigned negative infinite likelihoods from the current minibatch. In case this would remove more than $75\%$ of the minibatch, we skip the entire minibatch. These methods are only meant to ensure numerical stability, no specific training stability methods like gradient norm clipping are used.

\subsection{Outlier Loss Hyperparameters}
As described in the main manuscript, our outlier loss is:
\begin{equation}
L_{\mathrm{o}}=-\lambda\cdot\log\left(\sigma\left(\frac{\log(p_{\mathrm{g}}(x_{\mathrm{g}})) - \log(p_{\mathrm{in}}(x_{\mathrm{g}}))}{T}\right)\right)=-\lambda\cdot\log\left(\frac{\sqrt [T]{p_{\mathrm{g}}(x_{\mathrm{g}})}}{\sqrt[T]{p_{\mathrm{in}}(x_{\mathrm{g}})} + \sqrt[T]{p_{\mathrm{g}}(x_{\mathrm{g}})}}\right),
\label{eq:negative-loss.supp}
\end{equation}
where $\sigma$ is the sigmoid function, $T$ a temperature and $\lambda$ is a weighting factor. Based on a brief manual search on CIFAR10, we use $T=1000$  $\lambda=6000$ as they had the highest train set anomaly detection performance while retaining stable training. PixelCNN++ training works well with the same exact values, validating the choice.

\subsection{Evaluation Details}
We use the likelihoods computed on noise-free inputs for anomaly detection with Glow networks. In practice, this means not adding dequantization noise and instead adding a constant, namely half of the dequantization interval. We found this to yield slightly better anomaly detection performance in preliminary experiments. Noise-free inputs are only used during evaluation for the anomaly detection performance. Training is done adding the standard dequantization noise introduced as in \citep{Theis2015d}. The BPD numbers reported in Section \ref{sec:bpds} and Table \ref{tab:bpds} also are obtained using single samples with the standard dequantization noise.

We clip log likelihoods from below by a very small number. When computing the log likelihoods from networks trained with outlier loss, negative infinities are to be expected, see Section \ref{subsec:numerical}. To include these inputs in the AUC computation, we set non-finite log-likelihoods to a very small constant ($-3000000$) before computing any log-likelihood difference.

\subsection{Computing Infrastructure}
All experiments were computed on single GPUs. Runtimes vary between $2$ to $8$ days on Nvidia Geforce RTX 2080 depending on the experiment setting (outlier loss or not, supervised/unsupervised).
\section{Datasets}\label{sec:datasets}
\subsection{Dataset Splits}
We use the pre-defined train/test folds on CIFAR10, CIFAR100, SVHN, Fashion-MNIST and MNIST and only train on the training fold. All results are reported on the test folds. For CelebA, we only use the first $60000$ images for faster computations. We use a $1$ million random subset of 80 Million Tiny Images in all of our experiments. For the Fashion-MNIST/MNIST experiments, we create a greyscaled Tiny dataset from the rgb data as $x=r\cdot0.2989 + g\cdot 0.5870 + b\cdot0.1140$.

\subsection{MRI Dataset}
For the MRI BRATS dataset, we also use the official train/test split. The dataset was introduced to verify the log likelihood difference on a data from a slightly different domain (medical imaging). Since our outliers defined as different modalities are not defined by the object type, we did not expect last-scale likelihood contributions $c_3(x)$ to perform as good as on the object recognition datasets. However, they still outperform raw likelihoods, by 59.2\% to 53.3\%.

\section{Replication of \citep{Serra2020Input}}\label{sec:replicate-serra}

The anomaly detection results from \citep{Serra2020Input} were obtained using the training folds of the in-distribution datasets, preventing a fair comparison to our results. In written communication with~\citet{Serra2020Input}, they explained to us that the AUROC-results reported in their paper compare in-distribution training-fold examples with out-of-distribution test-fold examples. This makes a fair comparison to our results and other works impossible. In contrast and in line with standard practice, our results were obtained using the test folds of the in-distribution datasets. Unfortunately, \citet{Serra2020Input} are unable to provide their training code and models at the current time, so we cannot recompute their anomaly detection performance for the test fold of the in-distribution datasets. We also confirmed that depending on the training setting, the anomaly-detection AUROC values can differ substantially between the training and test fold of the in-distribution dataset.

In any case, we provide supplementary code to reproduce the method of\citep{Serra2020Input} to the best of our understanding. Using a publicly available pretrained Glow-model\footnote{\url{https://github.com/y0ast/Glow-PyTorch}}, we find anomaly detection performance results similar to the ones we report for PNG as a general-distribution model.

\section{Further Quantitative Results}\label{sec:quantitative-results}
\subsection{Maximum Likelihood Performance}
\label{sec:bpds}
The maximum-likelihood-performance of our finetuned Glow networks are similar to the performance reported for from-scratch training in the original Glow paper \cite{NIPS2018_8224}.  We show the bits-per-dimension values obtained using single dequantization samples in Table \ref{tab:bpds}. Note that the Glow model trained on 80 Million Tiny Images already reaches bits per dimensions on CIFAR10 and CIFAR100 close to the Glow models trained on the actual dataset (CIFAR10/CIFAR100), in line with our view that the bits per dimension are dominated by the domain prior (results also do not substantially change when including or excluding CIFAR-images from Tiny).

Our Glow-model architecture was chosen from a reimplementation of \citet{nalisnick2018do} (see Section \ref{subsec:full-glow-architecture}), in order to facilitate comparison of our results to other anomaly detection works. In future work, evaluating anomaly detection performance of our method with newer types of normalizing flows could be interesting.

\begin{table}%[htbp]
\setlength{\tabcolsep}{4pt}
\caption{Maximum likelihood performance in bits per dimension. Results obtained using single samples of uniform dequantization noise. Tiny is the Glow network trained on 80 Million Tiny Images. Retr refers to from-scratch training on the in-distribution dataset, Finet refers to finetuning aforementeioned Glow network trained on 80 Million Tiny Images. Note the original Glow paper \cite{NIPS2018_8224} reached 3.35 bpd on CIFAR-10 with multi-GPU training. The Glow network and training setup we use is optimized for single-GPU training and not for maximum performance. The public implementation we originally based our implementation on (and uses the explicit conditioning step discussed in \ref{subsec:independent-z}) reaches $3.39$ bpd on CIFAR10.
}

\begin{center}
\begin{tabular}{cccccc}
 \hline
 \textbf{In-dist} & \textbf{Tiny} & \textbf{Retr} & \textbf{Finet} \\
\hline
\textbf{SVHN    } & 2.34 & 2.07 & 2.06 \\
\textbf{CIFAR10 } & 3.41 & 3.40 & 3.36 \\
\textbf{CIFAR100} & 3.43 & 3.43 & 3.39 \\
\hline
\end{tabular}
\end{center}

\label{tab:bpds}\vspace{-0.6cm}
\end{table}

\subsection{Finetuning}\label{subsec:finetuning}

\begin{figure}[t!]
    \centering
    \includegraphics[width=0.8\textwidth]{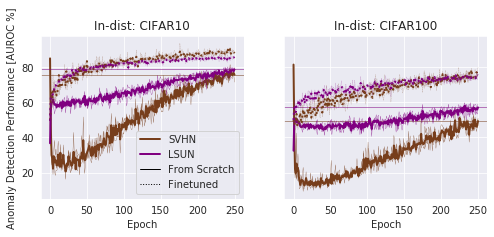}
    \caption{Training Curves Anomaly Detection From Scratch vs. Finetuned.  Conventions as in Fig. \ref{fig:training-curves-bpd}.  Glow networks are trained without any outlier loss. AUROC refers to AUROC computed from our log-likelihood ratio metric using another Glow-network trained on 80 Million Tiny Images. Note that the finetuned Glow networks outperform the final from-scratch trained Glow networks after less than $20\%$ of the training epochs.  Note that due to different evaluation (not noise-free) and different subsets used for intermediate results, results in this figures vary from final results in result tables.}
    \label{fig:training_curves_auc}
\end{figure}

Training Glow networks on an in-distribution dataset by finetuning a Glow network trained on  Tiny substantially speeds up the training progress over training from scratch. As can be seen in Figures \ref{fig:training-curves-bpd} and \ref{fig:training_curves_auc}, the Glow networks reach better results after less training epochs for both maximum-likelihood performance and anomaly-detection performance. The improvements are strongest for CIFAR100 and weakest for SVHN, in line with CIFAR100 being the most diverse dataset and most similar to 80 Million Tiny Images.

\begin{minipage}{\textwidth}
\begin{minipage}[b]{0.45\textwidth}
\centering
\includegraphics[width=\linewidth]{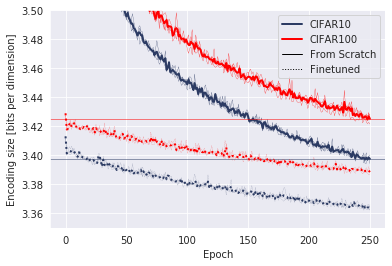}
\captionof*{figure}{\textbf{CIFAR10/100}}
\end{minipage}
\hspace{0.02cm}
\begin{minipage}[b]{0.45\textwidth}
\centering
\includegraphics[width=\linewidth]{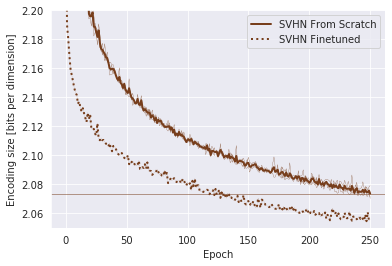}
  \captionof*{figure}{\textbf{SVHN}}
\end{minipage}
\captionof{figure}{Training Curves CIFAR10/100 and SVHN From Scratch vs. Finetuned. Transparent, thin lines indicate single-seed runs, solid, think lines indicate means over these runs. Solid horizontal lines indicate final mean performance of from-scratch trained models. Note that (i) finetuned Glow networks are better in each epoch; (ii) for CIFAR10/100 the finetuned Glow networks outperform the final from-scratch trained Glow networks after less than $20\%$ of the training epochs and (iii) for SVHN, the finetuned Glow network outperforms the final from-scratch-trained Glow network after about 50\% of the training epochs. }
\label{fig:training-curves-bpd}
\end{minipage}

\begin{figure}
    \centering
    \includegraphics[width=0.8\textwidth]{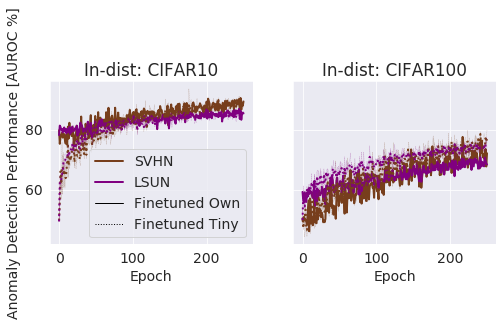}
    \caption{Training Curves Anomaly Detection Finetuned from Own model vs Finetuned from Tiny.  Conventions as in Fig. \ref{fig:training_curves_auc}.  Finetuned from own model is the same as simply training twice as long on the in-distribution. At the end of training for CIFAR100, the model finetuned from Tiny still performs \textasciitilde 6\% better on anomaly detection.}
    \label{fig:training_curves_auc_own_tiny}
\end{figure}

There are are still gains on log-likelihood ratio based anomaly detection for CIFAR100 when comparing finetuning a Glow a Glow network trained on Tiny  to finetuning a Glow network already trained on CIFAR100 (or in other words, training the Glow Network from scratch on CIFAR100 for twice the number of epochs), see Figure \ref{fig:training_curves_auc_own_tiny}. This is likely because the exact Tiny-model be used as the general distribution model later on, validating more similar models better cancel the model bias.

\subsection{Result Variance across Seeds}\label{subsec:seed-variance}
\begin{figure*}
    \centering
    \includegraphics[width=\linewidth]{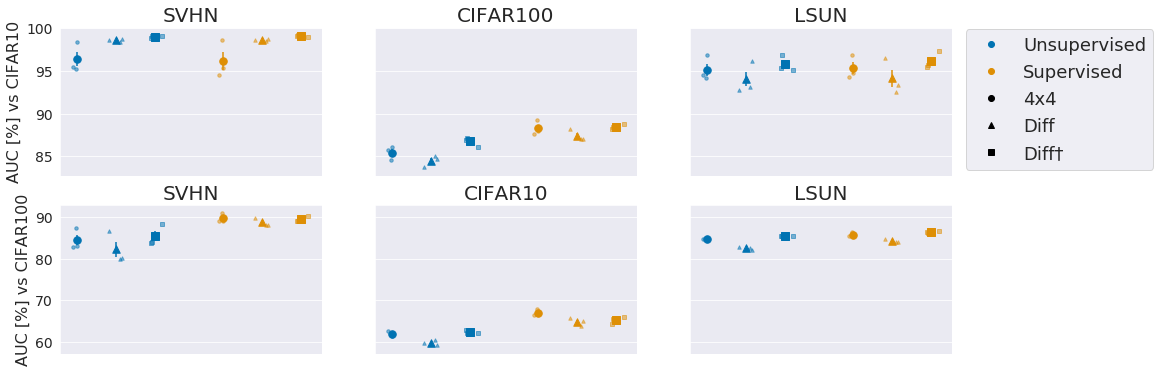}
    \caption{Graphical Overview over Anomaly Detection Results. Markers indicate mean result over three seeds, error bars indicate standard error of that mean. Type of marker indicates type of anomaly metric (defined as before and as in Table \ref{tab:overview-results-with-variance}). Color indicates supervised or unsupervised setting. Rows are in-distribution dataset and columns are OOD datasets. Supervised setting outperforms unsupervised setting, especially on CIFAR10 vs. CIFAR100 and vice versa. Using a general-distribution model trained with outlier loss on the in-distribution ({Diff\dag}) always outperforms general-distribution model trained without outlier loss (Diff). Relative performance of final-scale method ($4\times 4$) compared with log-likelihood-difference methods (Diff and Diff\dag) varies between dataset pairs.}
    \label{fig:results-scatterplot}
\end{figure*}

We present the original results including standard deviation in Table \ref{tab:overview-results-with-variance} and provide a graphical overview over our per-seed anomaly detection results in Figure \ref{fig:results-scatterplot}. Results are relatively stable across seeds.

\subsection{Additional OOD Datasets}\label{subsec:more-ood}
We report results on CelebA \cite{Liu_Celeba} and Tiny Imagenet \footnote{\url{https://tiny-imagenet.herokuapp.com}} as additional out-of-distribution (OOD) datasets in Table \ref{tab:other-ood-results}.

\begin{table*}%[htbp]
\caption{Anomaly detection performance summary (AUC in $\%$). Values in parentheses are standard deviation across 3 seeds. The new term \textbf{Diff\dag} means to use in-distribution samples as the outliers to train Tiny-Glow, see Sec. 5.4 in main paper.}

\begin{center}
\begin{tabular}{cc|ccc|ccc}
 \hline
 \textbf{}& \textbf{Setting} & \multicolumn{3}{c}{\textbf{Unsupervised}}& \multicolumn{3}{c}{\textbf{Supervised}} \\
 \hline
 \textbf{In-dist} & \textbf{Out-dist} & \textbf{4x4} & \textbf{Diff} & \textbf{Diff}\dag & \textbf{4x4} & \textbf{Diff} & \textbf{Diff}\dag   \\
\hline
\multirow{3}{*}{\textbf{CIFAR10  }} & SVHN      & 96.4 (1.4) & 98.6 (0.1) & 99.0 (0.1) & 96.1 (1.7) & 98.6 (0.1) & 99.1 (0.1)\\ 
                                    & CIFAR100  & 85.4 (0.7) & 84.5 (0.6) & 86.8 (0.5)  & 88.3 (0.7) & 87.4 (0.5) & 88.5 (0.3)\\ 
                                    & LSUN      & 95.1 (1.2) & 94.1 (1.5) & 95.8 (0.8) & 95.3 (1.1) & 94.1 (1.7) & 96.2 (0.9)\\ 
\hline 
&Mean                                           & 92.3 (1.0) & 92.4 (0.6) & 93.8 (0.4) &  93.3 (1.2) & 93.4 (0.8) & 94.6 (0.4) \\ 
\hline 
\multirow{3}{*}{\textbf{CIFAR100 }} & SVHN      & 84.5 (2.1) & 82.2 (3.1) & 85.4 (2.1)   & 89.6 (1.0) & 88.6 (0.8) & 89.4 (0.7) \\ 
                                    & CIFAR10   & 61.9 (0.5) & 59.8 (0.5) & 62.5 (0.3)   & 67.0 (0.6) & 64.9 (0.8) & 65.3 (0.7)\\ 
                                    & LSUN      & 84.6 (0.1) & 82.4 (0.3) & 85.4 (0.1)    & 85.7 (0.4) & 84.3 (0.3) & 86.3 (0.2) \\ 
\hline 
&Mean                                           & 77.0 (0.7) & 74.8 (1.1) & 77.8 (0.6) &   80.8 (0.7) & 79.3 (0.6) & 80.3 (0.5) \\ 
\hline 
\hline 
&Mean                                           & 84.7 (0.3) & 83.6 (0.3) & 85.8 (0.2)    & 87.0 (0.9) & 86.3 (0.7) & 87.5 (0.4) \\ 
\hline 
\hline 
\end{tabular}
\end{center}

\label{tab:overview-results-with-variance}
\end{table*}

\begin{table*}%[htbp]
\caption{Anomaly detection performance for additional OOD datasets CelebA and Tiny-Imagenet. Conventions as in Table main manuscript.}

\begin{center}
\begin{tabular}{cc|ccc|ccc}
 \hline
 \textbf{}& \textbf{Setting} & \multicolumn{3}{c}{\textbf{Unsupervised}}& \multicolumn{3}{c}{\textbf{Supervised}} \\
 \hline
 \textbf{In-dist} & \textbf{Out-dist} & \textbf{Raw [4x4]} & \textbf{Diff} & \textbf{Diff}\dag & \textbf{Raw [4x4]} & \textbf{Diff} & \textbf{Diff}\dag\\
\hline
\multirow{2}{*}{\textbf{CIFAR10  }}     & CelebA    & 96.6 (1.2) & 96.1 (1.2) & 97.6 (0.5) & 96.6 (1.9) & 96.2 (2.1) & 97.8 (1.0)  \\ 
                                    & Tiny-Imagenet & 90.7 (0.9) & 90.6 (0.7) & 92.1 (0.4)    & 91.1 (0.8) & 91.3 (0.9) & 92.7 (0.4)   \\ 
\hline 
\multirow{2}{*}{\textbf{CIFAR100 }}                                    & CelebA    & 80.9 (1.3) & 76.4 (2.4) & 80.4 (1.1)    & 81.9 (5.4) & 79.1 (7.2) & 81.7 (4.7)   \\ 
                                    & Tiny-Imagenet & 77.3 (0.5) & 77.5 (0.5) & 79.7 (0.3)    & 79.5 (0.4) & 79.7 (0.5) & 80.6 (0.5)   \\ 
\hline

\end{tabular}
\end{center}

\label{tab:other-ood-results}
\end{table*}

\subsection{Margin Loss vs. Outlier Loss}\label{subsec:loss-types}

\begin{figure}
    \centering
    \includegraphics[scale=0.5]{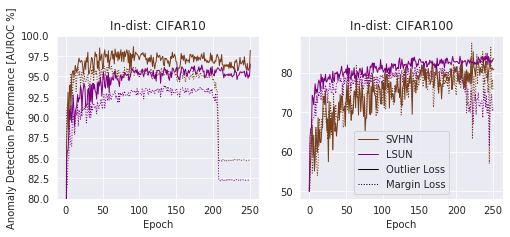}
    \caption{
    Training Curves Anomaly Detection Margin Loss vs Outlier Loss.
    AUROC refers to AUROC computed from our log-likelihood-difference metric using another Glow-network trained on 80 Million Tiny Images. Note Glow networks trained with margin loss experience substantial drops in anomaly detection performance in later stages of the training.
    }
    \label{fig:training_curves_margin_outlier}
\end{figure}

In our experiments, the margin-based loss introduced in \citep{hendrycks2018deep} is less stable than our outlier loss for longer training runs, see Figure \ref{fig:training_curves_margin_outlier}. Note that our results for the margin-based loss already substantially outperform the results reported for PixelCNN with a margin-based loss in \citep{hendrycks2018deep}.

\section{Qualitative Analyses}\label{sec:qualitative-results}
\begin{figure*}
    \centering
    \includegraphics[width=\linewidth]{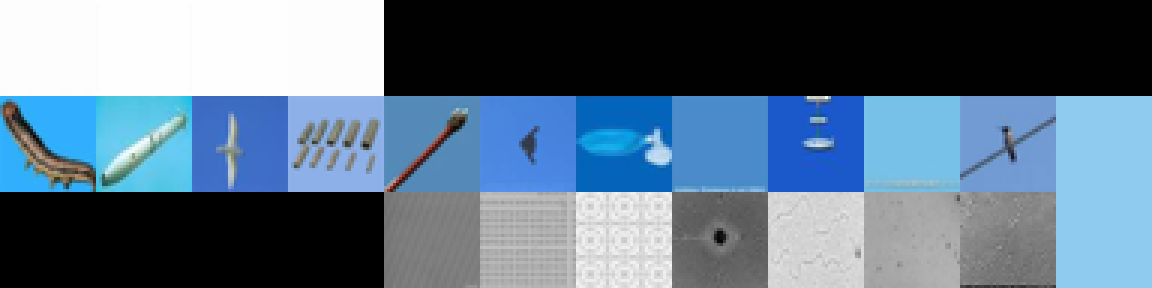}
    \caption{Most likely images from 80 Million Tiny Images for CIFAR10-Glow.
    12 highest-scoring images selected according to different metrics. First row: raw likelihood, second row: log-likelihood ratio to Tiny-Glow, third row: raw last-scale $z_3$ likelihood contribution. Note that constant images attain the highest raw likelihood, showing the effect of the natural-images domain prior on the raw likelihoods. The highest-scoring log-likelihood-ratio images show a bias towards blue images and some contain actual CIFAR10 objects, namely  birds and planes. Overall, the difference selects some correct images, but is still sensitive to surface features such as the global color. The last-scale results are harder to interpret, the more diverse images suggest it is slightly less affected by the domain prior of smoothness.}
    \label{fig:high-score-tiny}
\end{figure*}

\begin{figure*}
    \centering
    \includegraphics[width=\linewidth]{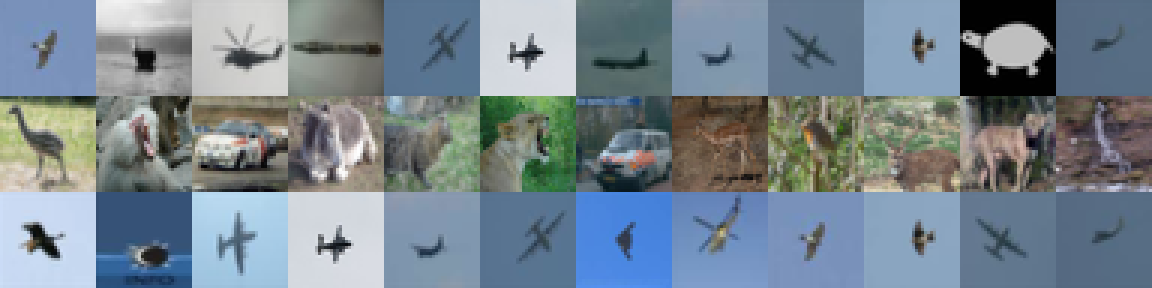}
    \caption{Most likely images from 80 Million Tiny Images for CIFAR10-Glow with the outlier loss. Conventions as in Figure \ref{fig:high-score-tiny-outlier-loss}. Now any of the three metrics lead to selecting mostly CIFAR10-like images, with the log-likelihood difference metric selecting more diverse images with less of a bias towards blueish images.}
    \label{fig:high-score-tiny-outlier-loss}
\end{figure*}

Our different metrics (raw likelihoods, log-likelihood ratios and last-scale likelihood contributions) result in qualitatively different highest-scoring images on 80 Million Tiny Images (see Fig. \ref{fig:high-score-tiny} and \ref{fig:high-score-tiny-outlier-loss}). We take a random 120000-images subset of 80 Million Tiny Images and use our Glow network trained on CIFAR10 either with or without outlier loss to compute the  metrics. Looking at the top 12 images per metric shows that using the log-likelihood ratio results in more reasonable images (closer to the inliers) than the raw likelihood, albeit mostly still simple images (see Fig.~\ref{fig:high-score-tiny}) and that using the Glow network trained with outlier loss results in more fitting images for all metrics (see Fig.~\ref{fig:high-score-tiny-outlier-loss}).

\clearpage

\section{Pure Discriminative Approach} \label{sec:binary-classifier}
As an additional baseline, we also evaluated using a purely discriminative approach. We trained a Wide-ResNet classifier to distinguish between the in-distribution and 80 Million Tiny-Images, without using any in-distribution labels. Concretely, we trained the classifier using samples of the in-distribution as the positive class and samples from 80 Million Tiny Images as the negative class in a normal supervised training setting. We use the training settings and architecture from a publicly available Wide-ResNet repository \footnote{\url{https://github.com/meliketoy/wide-resnet.pytorch}, with depth=28 and widen-factor=10}. After training, we use the prediction $p_{ResNet}(y_{indist}|x)$ as our anomaly metric. 

For CIFAR10/100 in-distribution, results show this baseline performs better for OOD dataset CIFAR100/10 (89\% and 70\% vs. 87\% and 63\% AUROC), similar for OOD dataset LSUN (93\% and 89\% vs. 96\% and 86\%) and worse for OOD dataset SVHN (93\% and 73\% vs. 99\% and 85\%) compared to our unsupervised generative methods (compare Table \ref{tab:binary-classifier} to unsupervised in \ref{tab:overview-results-with-variance}). Future work may further show what properties, advantages and disadvantages these different approaches have.

\begin{table}%[htbp]
\caption{Binary Classifier Anomaly Detection Results. Wide-ResNet classifier trained on 80 Million Tiny Images vs in-distribution as binary classification. We use $p_{ResNet}(y_{indist}|x)$ as our anomaly metric after training for the AUC computations. }
\begin{center}
\begin{tabular}{ccc}
 \hline
 \textbf{In-dist}
 & \textbf{OOD} & \textbf{AUC} \\
\hline
\textbf{CIFAR-10} 
 & SVHN      & 93  \\
 & CIFAR-100 & 89 \\
 & LSUN      & 93  \\
 \hline
\textbf{CIFAR-100} 
 & SVHN      & 73  \\
 & CIFAR-10 & 70 \\
 & LSUN      & 89  \\
 \hline
\textbf{SVHN} 
 & CIFAR-10  & 100 \\
 & CIFAR-100 & 100  \\
 & LSUN      & 100  \\
\hline
\end{tabular}
\label{tab:binary-classifier}
\end{center}
\end{table}

\medskip

\small

\end{document}